\documentclass{article}
\usepackage{ijcai19}

\usepackage{times}
\usepackage{soul}
\usepackage{url}
\usepackage[hidelinks]{hyperref}
\usepackage[utf8]{inputenc}
\usepackage[small]{caption}
\usepackage{graphicx}
\usepackage{amsmath}
\usepackage{booktabs}
\usepackage{physics}
\usepackage{amssymb}
\usepackage{mathrsfs}
\graphicspath{ {./figures/} }
\usepackage{comment}
\usepackage{dirtytalk}

\title{Widely Linear Complex-valued Autoencoder: Dealing with Noncircularity in Generative-Discriminative Models}

\author{
Zeyang Yu\footnote{Zeyang Yu}\and
Shengxi Li\And
Danilo Mandic \\
\affiliations
Department of Electrical and Electronic Engineering, Imperial College London, UK\\
\emails
\{z.yu17, shengxi.li17, d.mandic\}@imperial.ac.uk
}

\begin{document}

\maketitle

\begin{abstract}
We propose a new structure for the complex-valued autoencoder by introducing additional degrees of freedom into its design through a widely linear (WL) transform. The corresponding widely linear backpropagation algorithm is also developed using the $\mathbb{CR}$ calculus, to unify the gradient calculation of the cost function and the underlying WL model. More specifically, all the existing  complex-valued autoencoders employ the strictly linear transform, which is optimal only when the complex-valued outputs of each network layer are independent of the conjugate of the inputs. In addition, the widely linear model which underpins our work allows us to consider all the second-order statistics of inputs. This provides more freedom in the design and enhanced optimization opportunities, as compared to the state-of-the-art. Furthermore, we show that the most widely adopted cost function, i.e., the mean squared error, is not best suited for the complex domain, as it is a real quantity with a single degree of freedom, while both the phase and the amplitude information need to be optimized. To resolve this issue, we design a new cost function, which is capable of controlling the balance between the phase and the amplitude contribution to the solution. The experimental results verify the superior performance of the proposed autoencoder together with the new cost function, especially for the imaging scenarios where the phase preserves extensive information on edges and shapes.
\end{abstract}

\section{Introduction}
With the significant improvement of computational power and the exponential growth of data, deep learning has become the most rapidly growing area in the field of artificial intelligence. Numerous techniques have been proposed in this area, which enable a deep neural network model to even surpass human-level performance. Driven by these outstanding opportunities, extensive research has been conducted on the use of deep learning for image and voice recognition.

An autoencoder is an unsupervised deep learning structure, proposed by Hinton \textit{et al.}~\cite{hinton2006fast,hinton2006reducing}. It automatically learns to compress the original high-dimensional data and extract meaningful features, which can then be used as an input for other models. Essentially, the autoencoder is a neural network which can be divided into two parts - an encoder and a decoder. The encoder maps the original data into a low-dimensional representation, whilst the decoder learns to reconstruct the original data from this low-dimensional representation. This is one of the most widely applied deep learning algorithms for dimensionality reduction and feature extraction.

Recently, complex-valued neural networks have received increasing attention due to their potential for easier optimization, faster learning and robustness, compared to real-valued ones~\cite{aizenberg2011complex,hirose2012complex,trabelsi2017deep}. Applications include radar image processing, antenna design, and forecasting in smart grid, to mention but a few ~\cite{hirose2013complex}. In this context, Arjovsky \textit{et al.}~\shortcite{arjovsky2016unitary} showed that using a complex representation in recurrent neural networks can increase the representation capacity. Călin-Adrian Popa~\shortcite{popa2017complex} verified that the complex-valued convolutional neural network (CVCNN) outperforms the real-valued one, with the biggest improvement in performance for the bigger kernel sizes. In terms of autoencoders, Ryusuke Hata \textit{et al.}~\shortcite{hata2016multi} illustrated that a complex-valued autoencoder can extract better features than the real-valued one. However, all the existing complex-valued neural networks are designed using the strictly linear transform, which assumes the complex outputs of each layer are independent of the conjugate parts of the inputs. This assumption limits the structure of the covariance matrix of the output of the complex linear transform, restricts the number of degrees of freedom, and thus may possibly lead to sub-optimum or unstable optimization process.

In this paper, we propose a new structure for the complex-valued autoencoder, referred to as the widely linear complex-valued autoencoder (WLCAE). The proposed autoencoder makes use of the widely linear transform~\cite{mandic2009complex} for the linear part of the autoencoder, which provides more degrees of freedom in the analysis and enhanced performance. To optimize the parameters in such autoencoder, we derive the corresponding widely linear backpropagation algorithm using the $\mathbb{CR}$ calculus~\cite{mandic2009complex,kreutz2009complex}. In order to further improve the performance of the proposed autoencoder in the complex domain, instead of the standard mean squared error, a phase-magnitude cost function is also proposed to improve the performance of the complex-valued autoencoder.

\section{Preliminaries}
\subsection{Widely Linear Transform}
In a real-valued autoencoder, the linear transform of a single layer is defined as $\textbf{z} = \textbf{Wa}$ where $\textbf{a}$ is the input to this layer, $\textbf{W}$ is the transform matrix, and $\textbf{z}$ is the output of the linear transform. By applying this linear transform, the features from the previous layer are combined according to specific weighting, which to a great extend determines the behavior of the neural network as a whole. However, in the complex domain, it has recently been recognized that there are two types of linear transforms - strictly linear transform and widely linear transform.

In a way similar to the real-valued transform, with complex-valued $\textbf{z}, \textbf{W}$ and $\textbf{a}$, the strictly linear transform in the complex domain is defined as
\begin{equation}
    \textbf{z} = \textbf{Wa}
\end{equation}

The so called \say{augmented representation} of the strictly linear transform then clearly shows the lack of its degrees of freedom, as two of the block diagonal matrices below are zero, that is 
\begin{equation} \label{eq:strictly}
    \textbf{\underline{z}} = 
    \begin{bmatrix}
        \textbf{z} \\
        \textbf{z}^{*}
    \end{bmatrix} =
    \begin{bmatrix}
        \textbf{W}, \textbf{0} \\
        \textbf{0}, \textbf{W}^{*}
    \end{bmatrix}
    \begin{bmatrix}
        \textbf{a} \\
        \textbf{a}^{*}
    \end{bmatrix}
\end{equation}

The widely linear transform in complex domain is defined as 
\begin{equation}
    \textbf{z} = \textbf{W}_1 \textbf{a} + \textbf{W}_2 \textbf{a}^{*}
\end{equation}
and its augmented representation is given by
\begin{equation}
    \textbf{\underline{z}} = 
    \begin{bmatrix}
        \textbf{z} \\
        \textbf{z}^{*}
    \end{bmatrix} =
    \begin{bmatrix}
        \textbf{W}_1, \textbf{W}_2 \\
        \textbf{W}_{2}^{\textbf{*}}, \textbf{W}_{1}^{*}
    \end{bmatrix}
    \begin{bmatrix}
        \textbf{a} \\
        \textbf{a}^{*}
    \end{bmatrix} =
    \textbf{\underline{W}}
    \textbf{\underline{a}}
\end{equation}

Compared to the strictly linear transform in (\ref{eq:strictly}), one additional transfer matrix is added to merge the information from the conjugate of the input.

Complex-valued $\textbf{z}$ and $\textbf{x}$ can also be represented via the following form
\begin{equation}
    \begin{split}
    \textbf{z} &= \textbf{m} + j\textbf{n} \\
    \textbf{x} &= \textbf{u} + j\textbf{v}
    \end{split}
\end{equation} 
where $\textbf{m}$ and $\textbf{u}$ are the real parts, $\textbf{n}$ and $\textbf{v}$ are the imaginary parts. 

Then, the augmented representation of both of these two complex linear transforms is equivalent to the following real linear transform
\begin{equation}
    \begin{bmatrix}
        \textbf{m} \\
        \textbf{n}
    \end{bmatrix} =
    \begin{bmatrix}
        \textbf{M}_{11}, \textbf{M}_{12} \\
        \textbf{M}_{21}, \textbf{M}_{22}
    \end{bmatrix}
    \begin{bmatrix}
        \textbf{u} \\
        \textbf{v}
    \end{bmatrix}
\end{equation}
where
\begin{equation} \label{eq:con}
    \begin{split}
        \textbf{W}_1 &= \frac{1}{2}[\textbf{M}_{11} + \textbf{M}_{22} + j(\textbf{M}_{21} - \textbf{M}_{12})] \\
        \textbf{W}_2 &= \frac{1}{2}[\textbf{M}_{11} - \textbf{M}_{22} + j(\textbf{M}_{21} + \textbf{M}_{12})]
    \end{split}
\end{equation}

By inspecting (\ref{eq:strictly}) and (\ref{eq:con}), the strictly linear transform assumes $\textbf{M}_{11} = \textbf{M}_{22}$ and $\textbf{M}_{21} = \textbf{M}_{12} = \textbf{0}$, which imposes a very stringent constraint during the optimization process~\cite{mandic2009duality}. On the other hand, the widely linear transform exhibits sufficient degrees of freedom to capture the full available second-order information, the so called augmented complex statistics.

\subsection{Generalized Derivatives in the complex domain}
Consider a complex-valued function $f(z) = u(x, y) + jv(x, y)$ where $z = x + jy$. This function is differentiable at $z$ if it simultaneously satisfies the Cauchy-Riemann Equations~\cite{riemann1851grundlagen}
\begin{equation}
    \begin{split}
        \pdv{u(x, y)}{x} &= \pdv{v(x, y)}{y} \\
        \pdv{v(x, y)}{x} &= -\pdv{u(x, y)}{y}
    \end{split}
\end{equation}

However, these conditions are too stringent for general optimization in autoencoders. In particular, it is obvious that any function that depends on both $z$ and $z^{*}$ does not satisfy the Cauchy-Riemann conditions, which means that it is not differentiable in the standard complex way. Unfortunately, the most widely applied cost function, the mean squared error, belongs to this type: it is a real function of complex variables which is defined via the multiplication between the residual and its conjugate. To address this issue, the $\mathbb{CR}$ calculus was proposed to calculate the gradient.

Specifically, $\mathbb{CR}$ calculus assumes that $z$ and $z^{*}$ are mutually independent. Therefore, we need to calculate two gradients termed the $\mathbb{R}$-derivative and the $\mathbb{R}^{*}$-derivative~\cite{mandic2009complex}. The $\mathbb{R}$-derivative is calculated by
\begin{equation}
    \pdv{f}{z}_{|z^{*}=const} = \frac{1}{2}(\pdv{f}{x} - j\pdv{f}{y})
\end{equation}
while the $\mathbb{R}^{*}$-derivative has the form
\begin{equation}
    \pdv{f}{z^{*}}_{|z=const} = \frac{1}{2}(\pdv{f}{x} + j\pdv{f}{y})
\end{equation}

Correspondingly, the chain rule is derived as
\begin{equation}
    \pdv{f(g(z))}{z} = \pdv{f}{g}\pdv{g}{z} + \pdv{f}{g^{*}}\pdv{g^{*}}{z}
\end{equation}

\section{Proposed Widely Linear Complex-valued Autoencoder}
\begin{figure}[t]
  \includegraphics[width=0.5\textwidth]{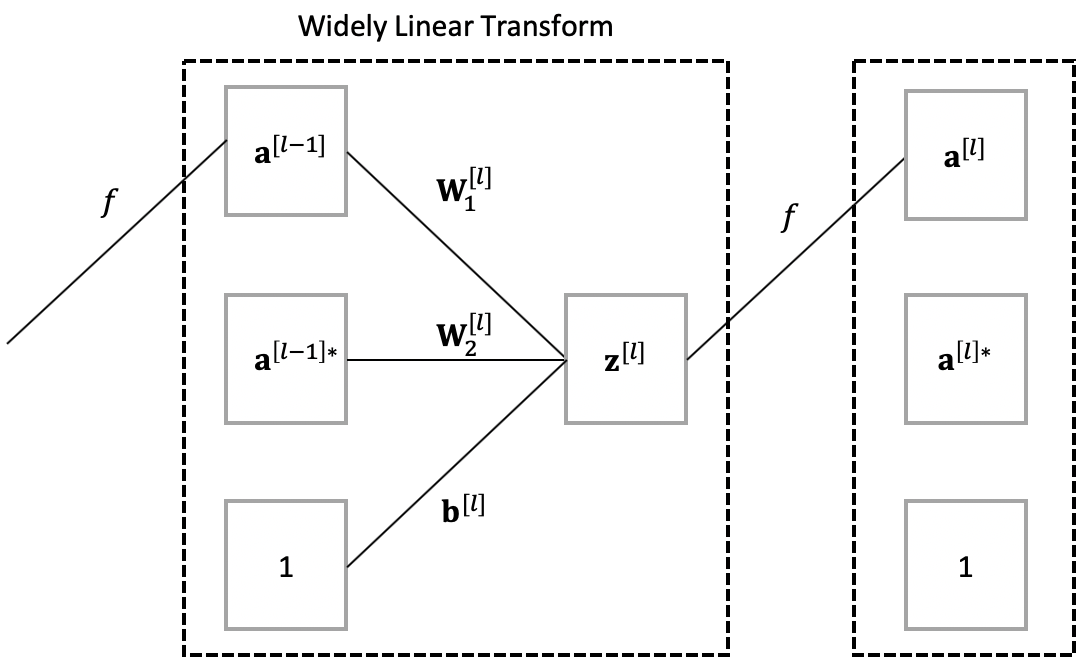}
  \caption{The structure of a single layer of the proposed widely linear complex-valued autoencoder.}
  \label{fig:autoencoder}
\end{figure}

The widely linear complex-valued autoencoder is an extension of the traditional complex-valued autoencoder which accounts for second-order data noncircularity (improperness). The structure of one single layer of the proposed widely linear network is shown in Figure~\ref{fig:autoencoder}. The widely linear complex-valued autoencoder includes two main building blocks: the widely linear transform component and an enhanced cost function which separates phase and amplitude for spectrum reconstruction. We first present the structure of the widely linear complex-valued autoencoder in Section \ref{WLCAE}, then we show in Section \ref{cost} that the phase is not well balanced when using the mean squared error cost function, which may lead to suboptimality of complex-valued autoencoders when using the mean squared error as the cost. To this end, we derive the novel phase-amplitude cost function on the basis of the popular mean squared error cost function in Section \ref{cost} to further improve the performance of the proposed autoencoder.

\subsection{Widely Linear Complex-Valued Autoencoder} \label{WLCAE}
Since the strictly linear transform limits the structure of the covariance matrix of the merged features, we propose to introduce a widely linear transform component into the complex-valued autoencoder. Let $\textbf{a}^{[0]}$ be the input vector, $\textbf{a}^{[l]}$ the output of the activation function, $\textbf{z}^{[l]}$ the result for the widely linear transform, $\textbf{W}^{[l]}_{1}, \textbf{W}^{[l]}_{2}$ the complex-valued weight matrices, and $\textbf{b}^{[l]}$ the bias vector. Moreover, $l \in [1, 2, \ldots, L]$ designates the index of each layer. The output vector of the activation function is then calculated as
\begin{equation}
    \begin{split}
    \textbf{z}^{[l]} &= \textbf{W}^{[l]}_{1}\textbf{a}^{[l-1]} + \textbf{W}^{[l]}_{2}\textbf{a}^{[l-1]*} + \textbf{b}^{[l]} \\
    \textbf{a}^{[l]} &= f(\textbf{z}^{[l]})
    \end{split}
\end{equation}
where $f(\cdot)$ is the activation function. The mean squared error cost function for one sample is given by
\begin{equation}
    J = \frac{1}{n}(\textbf{a}^{[L]} - \textbf{x})^{H}(\textbf{a}^{[L]} - \textbf{x})
\end{equation}
where $N$ is the dimension of $\textbf{a}^{[L]}$. 

Then, the widely linear backpropagation algorithm can be derived using the $\mathbb{CR}$ calculus as follows
\begin{equation}
    \begin{split}
    \pdv{J}{\textbf{z}^{[l]}} &= \pdv{J}{\textbf{a}^{[l]}}\pdv{\textbf{a}^{[l]}}{\textbf{z}^{[l]}} + \pdv{J}{\textbf{a}^{[l]*}}\Big[\pdv{\textbf{a}^{[l]}}{\textbf{z}^{[l]*}}\Big]^{*} \\
    \pdv{J}{\textbf{z}^{[l]*}} &= \pdv{J}{\textbf{a}^{[l]}}\pdv{\textbf{a}^{[l]}}{\textbf{z}^{[l]*}} + \pdv{J}{\textbf{a}^{[l]*}}\Big[\pdv{\textbf{a}^{[l]}}{\textbf{z}^{[l]}}\Big]^{*} \\
    \pdv{J}{\textbf{a}^{[l-1]}} &= \pdv{J}{\textbf{z}^{[l]}}\pdv{\textbf{z}^{[l]}}{\textbf{a}^{[l-1]}} + \pdv{J}{\textbf{z}^{[l]*}}\Big[\pdv{\textbf{z}^{[l]}}{\textbf{a}^{[l-1]*}}\Big]^{*} \\
    \pdv{J}{\textbf{a}^{[l-1]*}} &= \pdv{J}{\textbf{z}^{[l]}}\pdv{\textbf{z}^{[l]}}{\textbf{a}^{[l-1]*}} + \pdv{J}{\textbf{z}^{[l]*}}\Big[\pdv{\textbf{z}^{[l]}}{\textbf{a}^{[l-1]}}\Big]^{*}
    \end{split}
\end{equation}

When the activation function satisfies the Cauchy-Riemann equations, then $\pdv{\textbf{a}^{[l]}}{\textbf{z}^{[l]*}} = \textbf{0}$. In this case, the first two equations of the widely linear backpropagation can be simplified as follows
\begin{equation}
    \begin{split}
    \pdv{J}{\textbf{z}^{[l]}} &= \pdv{J}{\textbf{a}^{[l]}}\pdv{\textbf{a}^{[l]}}{\textbf{z}^{[l]}} \\
    \pdv{J}{\textbf{z}^{[l]*}} &= \pdv{J}{\textbf{a}^{[l]*}}\Big[\pdv{\textbf{a}^{[l]}}{\textbf{z}^{[l]}}\Big]^{*}
    \end{split}
\end{equation}

It should be pointed out that the main decent direction of the cost function in the complex domain is in the direction of the conjugate gradient~\cite{brandwood1983complex}. Since the conjugate gradients of the parameters with respect to $\textbf{z}^{[l]}$ all vanish, the gradient of the cost function with respect to the weight matrices and the bias vector can be expressed as
\begin{equation}
    \begin{split}
    \nabla_{\textbf{W}^{[l]}_{1}}J &= \pdv{J}{\textbf{z}^{[l]*}}\Big[\pdv{\textbf{z}^{[l]}}{\textbf{W}^{[l]}_{1}}\Big]^{*} \\
    \nabla_{\textbf{W}^{[l]}_{2}}J &= \pdv{J}{\textbf{z}^{[l]*}}\Big[\pdv{\textbf{z}^{[l]}}{\textbf{W}^{[l]}_{2}}\Big]^{*} \\
    \nabla_{\textbf{b}^{[l]}}J &= \pdv{J}{\textbf{z}^{[l]*}}\Big[\pdv{\textbf{z}^{[l]}}{\textbf{b}^{[l]}}\Big]^{*}
    \end{split}
\end{equation}

\begin{figure}[t]
  \includegraphics[width=0.5\textwidth]{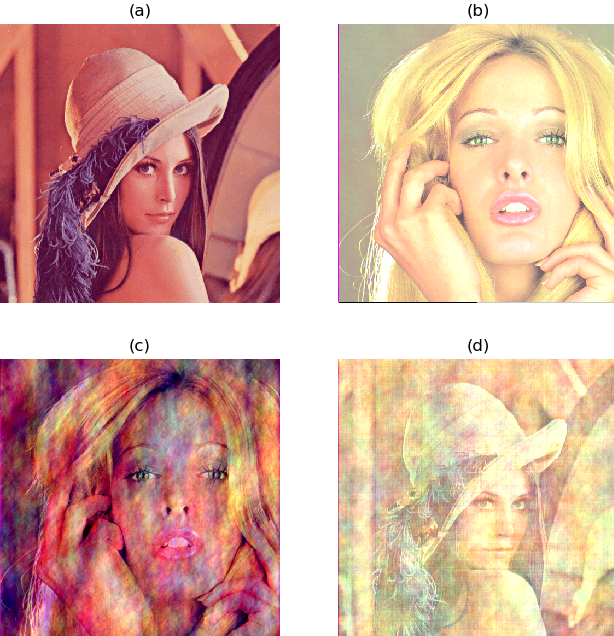}
  \caption{Effects of phase on images. (a) and (b): original images $I_{1}$ and $I_{2}$; (c) and (d): images $\hat{I}_{1}$ and $\hat{I}_{2}$ generated by swapping the phase spectra and keeping the original magnitude spectra.}
  \label{fig:phase}
\end{figure}

\begin{figure*}[t]
  \includegraphics[width=\textwidth]{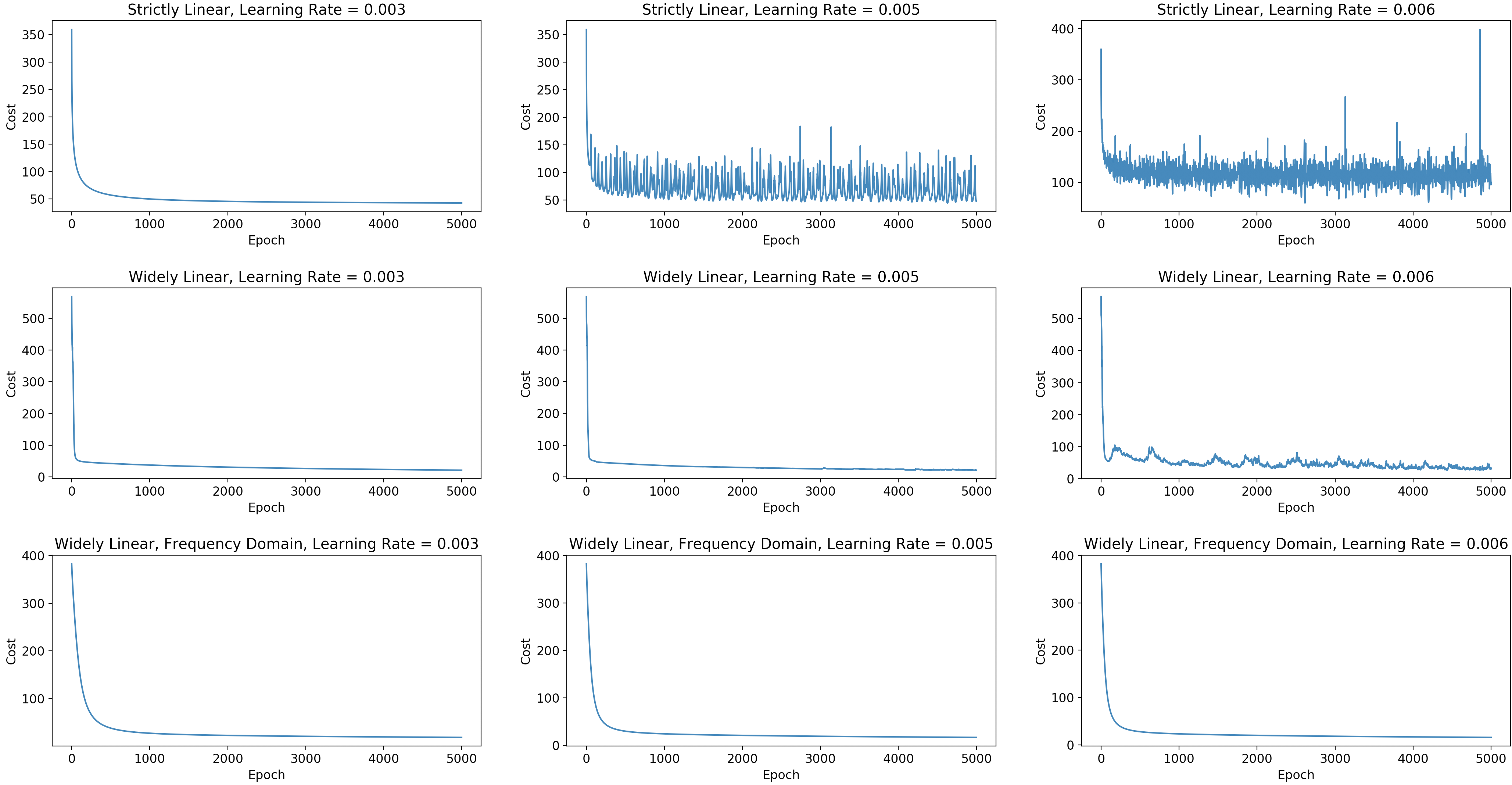}
  \caption{Training performance of the proposed widely linear autoencoder vs the standard one. Top: Strictly linear autoencoder, combining adjacent pixels into one complex number. Middle: Widely linear autoencoder, combining adjacent pixels into one complex number. Bottom: Widely linear autoencoder, Fourier transform. Learning rate from left to right: 0.003, 0.005 and 0.006.}
  \label{fig:stability}
\end{figure*}

\subsection{ Importance of phase information}\label{cost}
The complex number consists of the phase and the amplitude parts, where the phase of the frequency spectrum has been proven to play a more important role than the magnitude. Indeed, Oppenheim and Lim~\shortcite{oppenheim1981importance} verified that the information encoded in the magnitude of an image can be approximately recovered by the information encoded in its phase. For illustration, consider the surrogate images in Figure~\ref{fig:phase} with $I_{1}$ and $I_{2}$ denoting the two original images in the the top panel of Figure~\ref{fig:phase}. The 2D Fourier transform is first applied to these two images and their phase parts are swapped, which results in the new images at the bottom of Figure~\ref{fig:phase}, that is
\begin{equation}
    \begin{split}
        \hat{I}_{1} &= \mathscr{F}^{-1}(|\mathscr{F}(I_{1})|e^{j\angle\mathscr{F}(I_{2})}) \\
        \hat{I}_{2} &= \mathscr{F}^{-1}(|\mathscr{F}(I_{2})|e^{j\angle\mathscr{F}(I_{1})})
    \end{split}
\end{equation}

The example in Figure~\ref{fig:phase} makes it obvious that the phase preserves the information of edges and shapes while the amplitude preserves the information about pixel intensity. Inspired by this idea, we propose a new cost function for the complex-valued autoencoder in order to make it possible to balance the cost between amplitude and phase.

Mean squared error is one of the most widely used cost functions for the complex-valued autoencoder. However, this cost function suffers from the lack of ability to measure the difference in the frequency domain. As mentioned above, the importance of different frequency components in an image should not be treated on equal terms. In order to further enhance the performance of spectrum reconstruction, we propose to normalize the cost at each neuron by the amplitude of the corresponding input component, which from the above discussion is physically meaningful. To prevent the gradient from exploding, a lower bound, $\beta$, is added to the normalization factor, to yield a normalized cost for a single neuron in the form
\begin{equation} \label{cost_normlized}
    J_{i} = \frac{(a^{[L]}_{i} - x_{i})(a^{[L]}_{i} - x_{i})^{*}}{\max(x_{i}x_{i}^{*}, \beta)}
\end{equation}
where all the variables correspond to the neuron with index $i$.

Since the information contained in the phase is comparably important, we propose to separate the cost function into two costs - one of the amplitude and the other one of the phase. A similar practice has already yielded advantages in linear adaptive filtering~\cite{douglas2011least}. Let $x = A_{x}e^{j\theta_{x}}$ and $y = A_{y}e^{j\theta_{y}}$ be the values of two corresponding input and output neurons, where $A$ is the amplitude and $\theta$ is the phase. We start from the mean squared error of this pair of neurons which can be factorized as
\begin{equation}
    \begin{split}
        J &= (y - x)(y - x)^{*} \\
          &= (A_{y}e^{j\theta_{y}} - A_{x}e^{j\theta_{x}})(A_{y}e^{j\theta_{y}} - A_{x}e^{j\theta_{x}})^{*} \\
          &= (A_{y} - A_{x})^{2} + A_{y}A_{x}\big(2 - 2\cos(\theta_{y} - \theta_{x})\big)
    \end{split}
\end{equation}

The first term above is the cost of the amplitude, and the second term is the cost of the phase weighted by the amplitude of the input and output components. Then, we can adjust the weight of the second term to control the importance of phase during the reconstruction of the frequency spectrum. 

By combining the idea of normalizing the cost of reconstruction by amplitude and manipulating the cost of phase, we develop a new cost function which is specific for frequency spectrum reconstruction and has the form
\begin{equation} \label{cost_final}
    J = \frac{(\sqrt{yy^{*}} - \sqrt{xx^{*}})^{2} + \alpha(2\sqrt{yy^{*}xx^{*}} - xy^{*} - x^{*}y)}{\max(xx^{*}, \beta)}
\end{equation}

\section{Experiments}
The performances of a strictly linear and the proposed widely linear complex-valued autoencoder were evaluated on the benchmark MNIST database. For simplicity, both these types of autoencoder were implemented with only one hidden layer, and the performance was measured by peak signal-to-noise ratio (PSNR) of the reconstructed images.

\begin{table}[t]
    \centering
    \begin{tabular}{lrr}
    \toprule
    Input Data                  & Cost Function     & PSNR (dB)      \\
    \midrule
    Complex Pixels              & MSE               & 15.68          \\
    DFT                         & MSE               & 16.20          \\
    DFT                         & Normalized MSE    & 18.47          \\
    \bottomrule
    \end{tabular}
    \caption{PSNR values with different cost functions}
    \label{tab:reconstruction}
\end{table}

\begin{figure}[b]
  \includegraphics[width=0.5\textwidth]{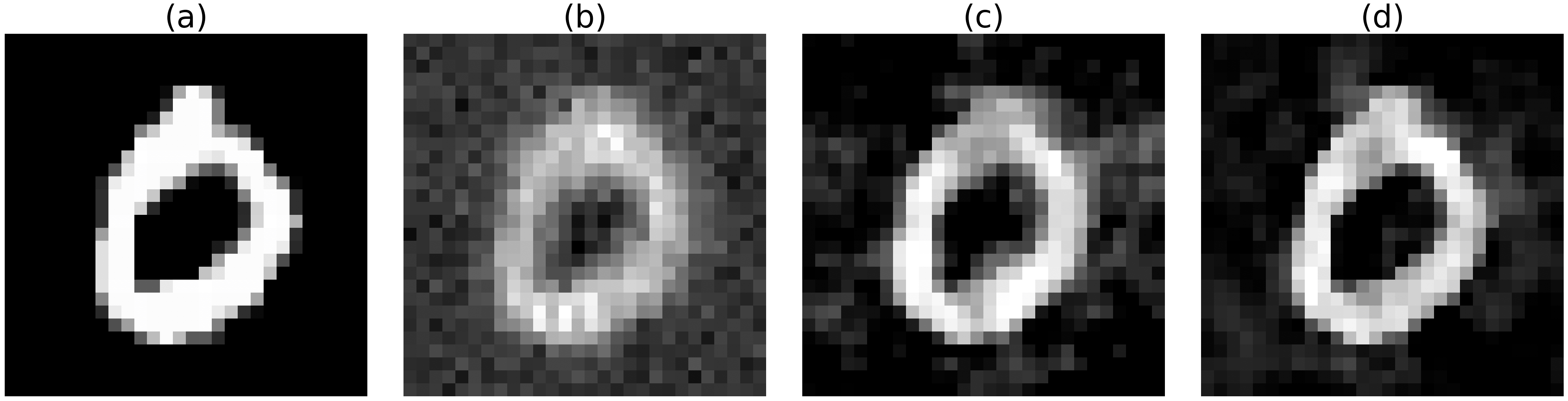}
  \caption{Examples of the reconstructed images. (a): Original image. (b): Combining adjacent pixels into one complex number as the input, mean squared error as the cost function. (c): Frequency spectrum as the input, mean squared error as the cost function. (d): Frequency spectrum as the input, mean squared error normalized by amplitude as the cost function.} 
  \label{fig:reconstruction}
\end{figure}

\subsection{Experimental Setup}
For the training data, we randomly selected 250 samples for each digital number, and then normalized the pixel values between 0 and 1. The complex image data can be generated in two ways. The first way is to combine adjacent pixels into one complex number, the most widely used way to generate complex data. The second way is to apply the 2D Fourier transform and discard the conjugate symmetric part of the data. The frequency spectrum was reconstructed via the output of the autoencoders and an inverse Fourier transform was applied to obtain the original image.

In our experiments, the strictly linear autoencoder had 196 hidden units while the widely linear autoencoder had 98 hidden units. As a result, both of these two autoencoders had the same number of parameters. For the activation function, we used the inverse tangent function $f(\cdot) = arctan(\cdot)$. For the initialization, we used Xavier initialization~\cite{glorot2010understanding} for both the real and imaginary parts, with the same random seed. 

To train the autoencoder, three cost functions were used - mean squared error, mean squared error normalized by amplitude shown in equation~\eqref{cost_normlized}, and the phase-amplitude cost function shown in equation~\eqref{cost_final}. Each autoencoder was trained over 5000 epochs. For simplicity, the lower bound of normalization $\beta$ was set to 0.1.

\subsection{Experimental Result}
Figure~\ref{fig:stability} shows the training process of each types of autoencoders with different learning rates. Observe that under the same type of the input, the proposed widely linear autoencoder was more stable than the strictly linear autoencoder. In addition, the use of the Fourier transform to generate complex data was able to stabilize the auto-encoder for large learning rates, which allows the acceleration of the training process.

Table \ref{tab:reconstruction} shows the effect on the PSNR of normalizing frequency components by their magnitude. We can see from the table that the PSNR was significantly reduced by using the normalized cost function. After applying the Fourier transform to the input data, the original data was transferred to an orthogonal representation, which is much easier to learn. However, the magnitude of different frequency components are no longer of the same scale. Normalizing the reconstruction error in the frequency domain with the magnitude was thus capable of significantly improving the performance of reconstruction of the original images. Figure~\ref{fig:reconstruction} shows the examples of the reconstructed images of these three algorithms.

\begin{figure}[t]
  \includegraphics[width=0.5\textwidth]{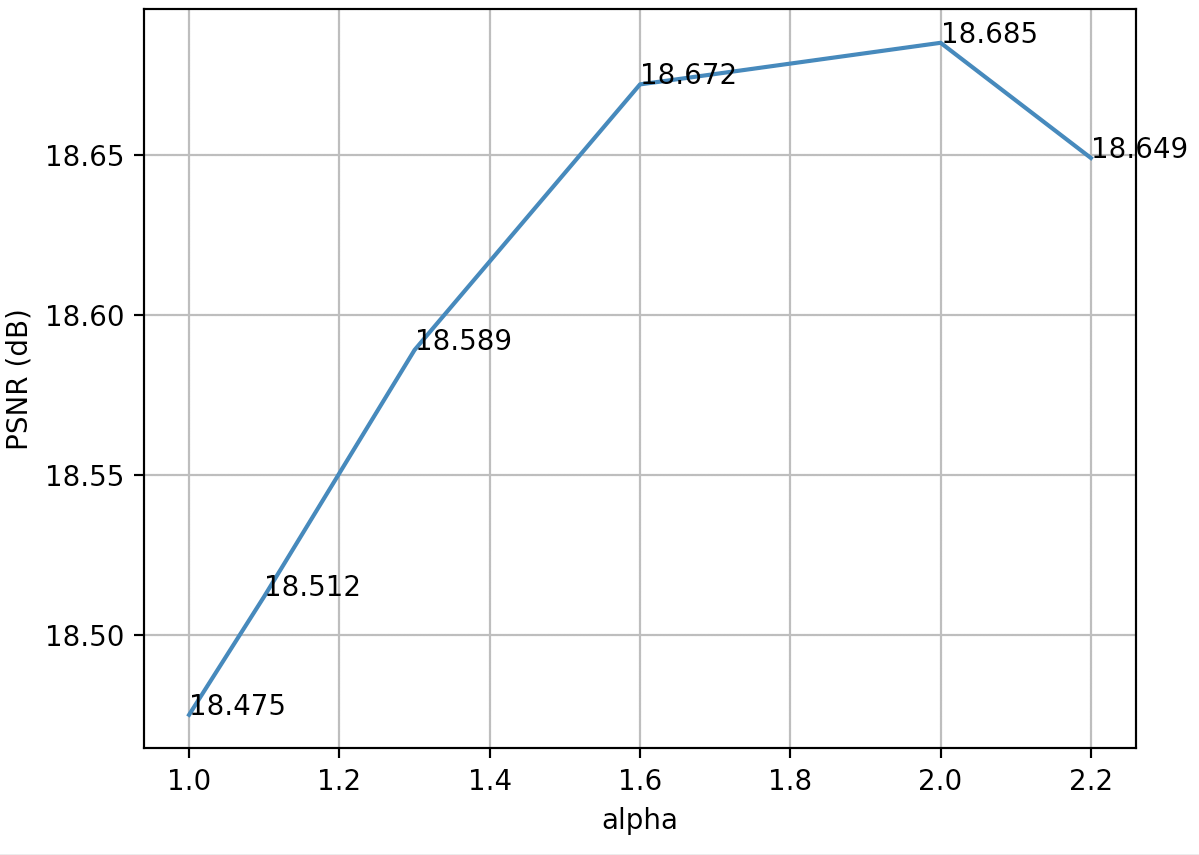}
  \caption{The effect of phase weighting, $\alpha$, on reconstruction.}
  \label{fig:final}
\end{figure}

Figure~{\ref{fig:final}} shows the advantageous effects of tuning separately the phase term in the cost function on PSNR. By increasing the weight parameter, $\alpha$, the performance was further improved.

\section{Conclusions}
We have proposed the widely linear complex-valued autoencoder to enhance the degrees of freedom in the design, and have introduced the phase-amplitude cost function, to math the requirements of spectrum reconstruction. Since the strictly linear transform in the complex domain cannot capture the whole second-order statistics as it uses only the covariance matrix, such optimization process is significantly unstable. By using the widely linear transform in the complex-valued autoencoder or deep neural network, we have shown that the stability can be significantly improved through the underlying augmented complex statistics. In addition, when applying the Fourier transform to the input data and training the autoencoder with the proposed phase-amplitude cost function, the reconstruction error has been shown to be significantly reduced. 

\bibliographystyle{ijcai19}
\bibliography{ijcai19}

\end{document}